\documentclass[sigconf,authorversion]{acmart}

\usepackage{booktabs} 
\usepackage[utf8]{inputenc} 
\usepackage[T1]{fontenc}    
\usepackage{url}            
\usepackage{booktabs}       
\usepackage{amsfonts}       
\usepackage{nicefrac}       
\usepackage{microtype}      
\usepackage{amsmath}
\usepackage{graphicx}
\usepackage{subcaption}
\usepackage{bm}
\usepackage[title]{appendix}
\usepackage{verbatim}

\usepackage{EvolvingAIcommenting}
\newif\ifcomments

\commentstrue

\ifcomments
\newcommand{\comments}[1]{#1}
\else
\newcommand{\comments}[1]{}
\fi

\setcopyright{rightsretained}

\acmDOI{10.1145/nnnnnnn.nnnnnnn}

\acmISBN{978-x-xxxx-xxxx-x/YY/MM}

\copyrightyear{2018} 
\acmYear{2018} 
\acmConference[GECCO '18 Companion]{Genetic and Evolutionary Computation Conference Companion}{July 15--19, 2018}{Kyoto, Japan}
\acmBooktitle{GECCO '18 Companion: Genetic and Evolutionary Computation Conference Companion, July 15--19, 2018, Kyoto, Japan}
\acmPrice{15.00}
\acmDOI{10.1145/3205651.3208236}
\acmISBN{978-1-4503-5764-7/18/07}

\begin{document}
\title{VINE: An Open Source Interactive Data Visualization Tool for Neuroevolution}

\author{Rui Wang, Jeff Clune, and Kenneth O. Stanley}
\affiliation{%
  \institution{Uber AI Labs}
  \city{San Francisco} 
  \state{CA 94103} 
  \postcode{94103}
}
\email{{ruiwang, jeffclune, kstanley}@uber.com}

\begin{abstract}
Recent advances in \emph{deep neuroevolution} have demonstrated that evolutionary algorithms, such as evolution strategies (ES) and genetic algorithms (GA), can scale to train deep
neural networks to solve difficult reinforcement learning (RL) problems. 
However, it remains a challenge to analyze and interpret the underlying
process of neuroevolution in such high dimensions.
To begin to address this challenge, this paper presents an interactive data visualization tool called VINE (Visual Inspector for NeuroEvolution) aimed at helping neuroevolution researchers and end-users better understand and explore this family of algorithms. VINE works seamlessly with a breadth of neuroevolution algorithms, including ES and GA, and addresses the difficulty of observing the underlying dynamics of the learning process through an interactive visualization of the evolving agent's behavior characterizations over generations. As neuroevolution scales to neural networks with millions or more connections, visualization tools like VINE that offer fresh insight into the underlying dynamics of 
evolution become increasingly valuable and important for inspiring new innovations and applications.
\end{abstract}

%



\maketitle


\section{Introduction}

\begin{figure}
  \centering
  \includegraphics[width=2.2in]{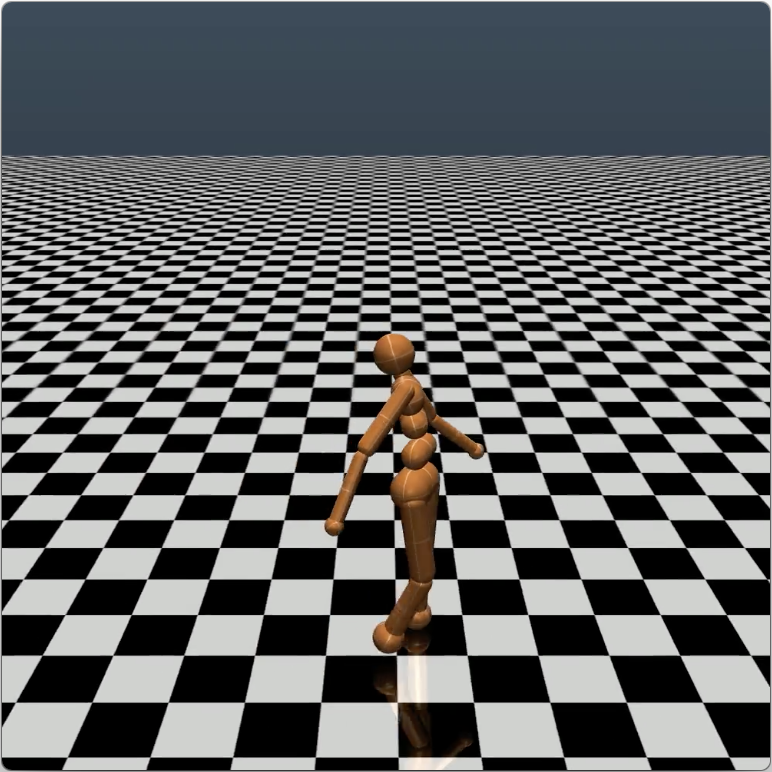}  
	\caption{\label{fig:esandga}\textbf{The Mujoco Humanoid Locomotion task.} This benchmark is the basis of a number of examples in this paper and can be solved by both the ES and GA approaches to neuroevolution.}
\end{figure}

Recent progress in \emph{deep neuroevolution} \citep{es,ken:oriley17,lehman:arxiv17sm,such:arxiv17,conti:arxiv17}  has shown that evolutionary algorithms, such as evolution strategies (ES) and genetic algorithms (GA), are capable of training deep neural networks \citep{goodfellow:deep} with millions or more parameters (weights) to solve difficult reinforcement learning (RL) problems. 
Figure \ref{fig:esandga} illustrates one such popular problem,
Mujoco Humanoid Locomotion, which both ES and GA solve effectively \cite{conti:arxiv17,such:arxiv17}.

While it is possible to probe the properties of such algorithms, such as in recent investigations into the relationship of ES to finite-difference gradient approximation \citep{lehman:arxiv17fd} and stochastic gradient descent \citep{zhang:arxiv17}, it is generally difficult to observe the underlying dynamics of the learning process in neuroevolution and neural network optimization. To address this gap and open up the process to observation, we introduce the \emph{Visual Inspector for NeuroEvolution} (VINE), an interactive data visualization tool aimed at helping those who are interested in neuroevolution to better understand and explore its behavior. The source code for VINE is available at \url{https://www.github.com/uber-common/deep-neuroevolution/tree/master/visual_inspector}. We hope this technology will inspire new understanding, innovations, and applications of neuroevolution in the future. 

VINE can illuminate both ES- and GA-style approaches. In this paper, we focus on visualizing the result of applying ES to the Mujoco Humanoid Locomotion \citep{todorov:mujoco,brockman:openaigym16} task from Figure \ref{fig:esandga}.

\section{Using VINE}


In the conventional application of the version of ES popularized by OpenAI \citep{es}, a group of neural networks called the pseudo-offspring cloud are optimized against an objective over generations. The parameters of each individual neural network in the cloud are generated by randomly perturbing the parameters of a single “parent” neural network. Each pseudo-offspring neural network is then evaluated against the objective: in the Humanoid Locomotion task, each pseudo-offspring neural network controls the movement of a robot, and earns a score called its fitness based on how well it walks. The ES constructs the next parent by aggregating the parameters of pseudo-offspring based on these fitness scores (almost like a sophisticated form of multi-parent crossover, and also reminiscent of stochastic finite differences). The cycle then repeats. The full details of this technique are formalized in \citep{es}.

To take advantage of VINE, behavior characterizations (BCs) \citep{lehman:ecj11} for each parent and all pseudo-offspring are recorded during evaluation. Here, a BC can be any property of the agent’s behavior when interacting with its environment. For example, in the Mujoco Humanoid Locomotion task we simply use the agent’s final \{\emph{x}, \emph{y}\} location as the BC, which indicates how far the agent has moved away from the origin and to what location.

\begin{figure}
  \centering
  \begin{subfigure}[t]{0.5\textwidth}
  		\centering
         \includegraphics[width=3.20in]{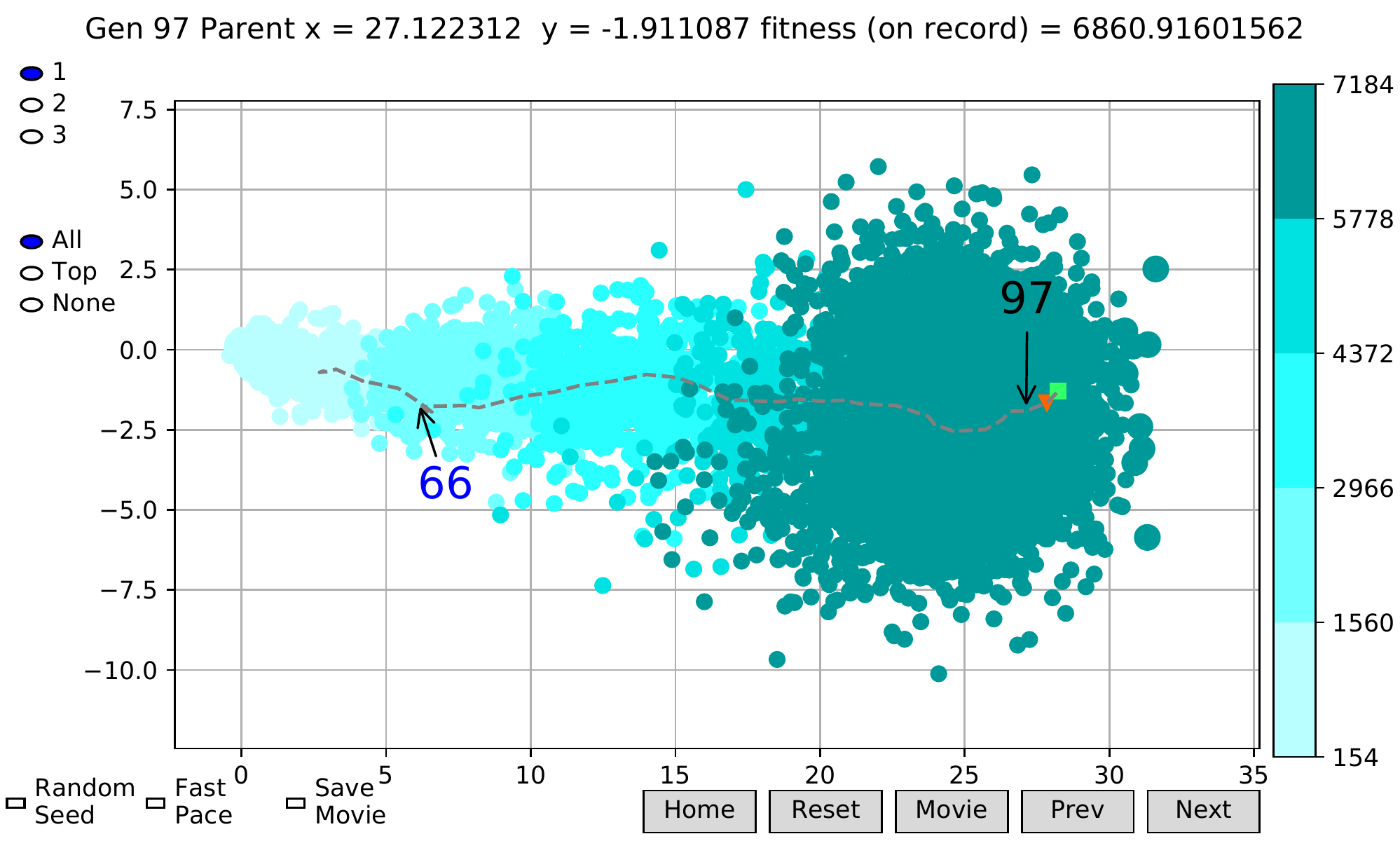}
        \caption{Cloud Plot}
    \end{subfigure}\\%
  \begin{subfigure}[t]{0.5\textwidth}
        \centering
  \includegraphics[width=3.20in]{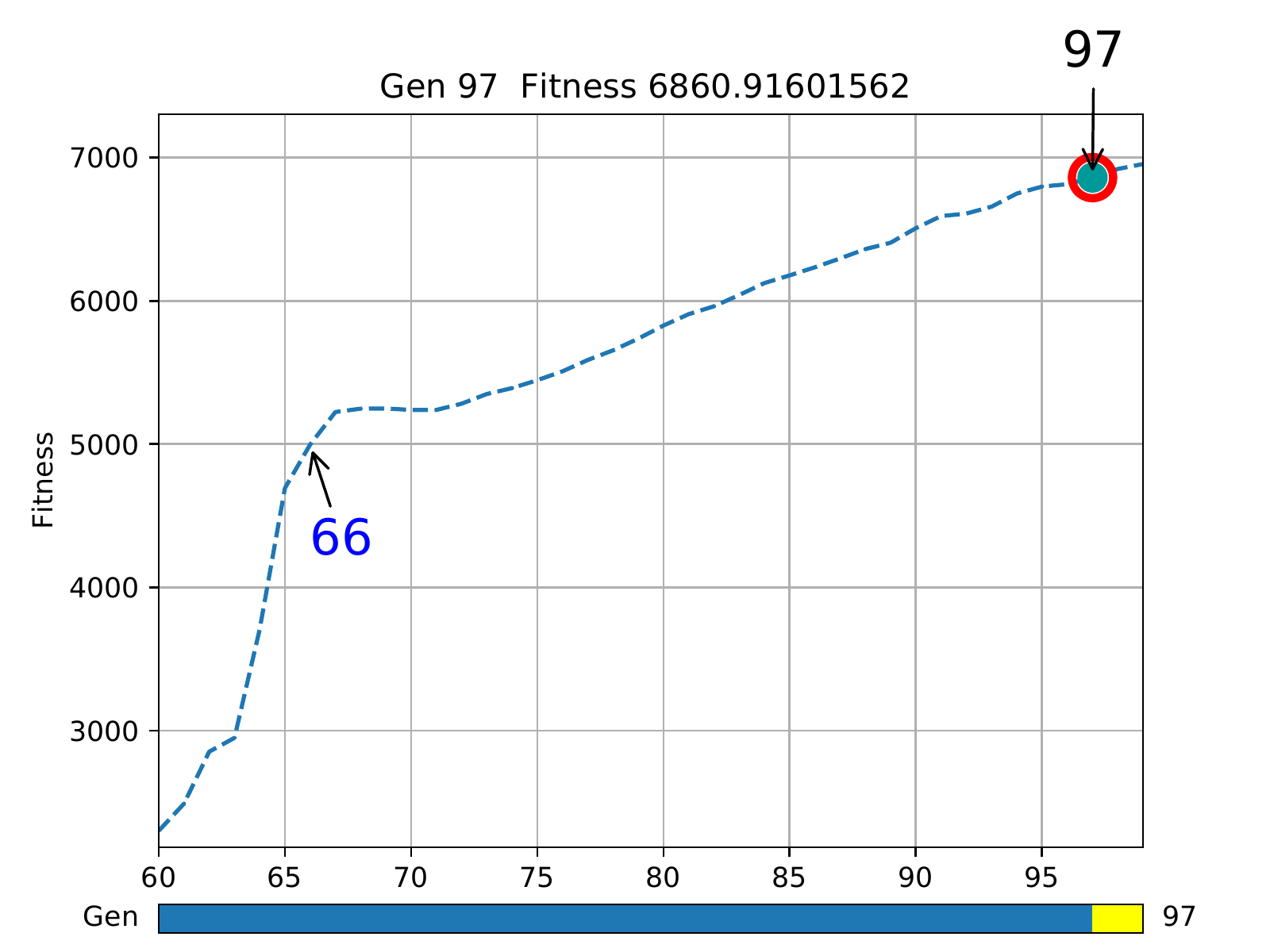}
  \caption{Fitness Plot}
  \end{subfigure}
  \centering
	\vspace{-0.1in}
	\caption{\label{fig:overview}\textbf{Examples of a pseudo-offspring cloud plot and a fitness plot.}}
\end{figure}

The visualization tool then maps parents and pseudo-offspring onto 2D planes according to their BCs.  For that purpose, it invokes a graphical user interface (GUI), whose major components consist of two types of interrelated plots: one or more pseudo-offspring cloud plots (on separate 2D planes), and one fitness plot. Illustrated in Figure \ref{fig:overview}, a pseudo-offspring cloud plot displays the BCs for the parent and pseudo-offspring in the cloud for every generation, while a fitness plot displays the parent’s fitness score curve as a key indicator of progress over generations. 

\begin{figure}
  \centering
  \begin{subfigure}[t]{0.5\textwidth}
  		\centering
         \includegraphics[width=3.00in]{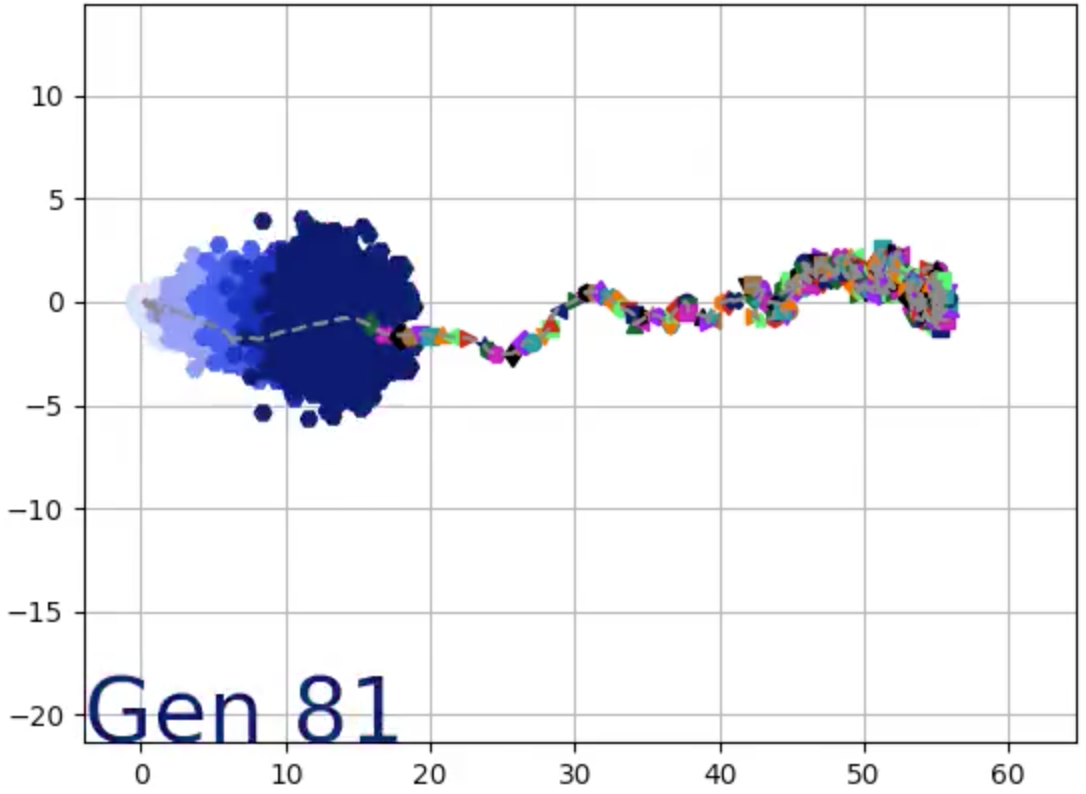}
        \caption{Video snapshot at Generation 81}
    \end{subfigure}\\%
  \begin{subfigure}[t]{0.5\textwidth}
  		\centering
         \includegraphics[width=3.00in]{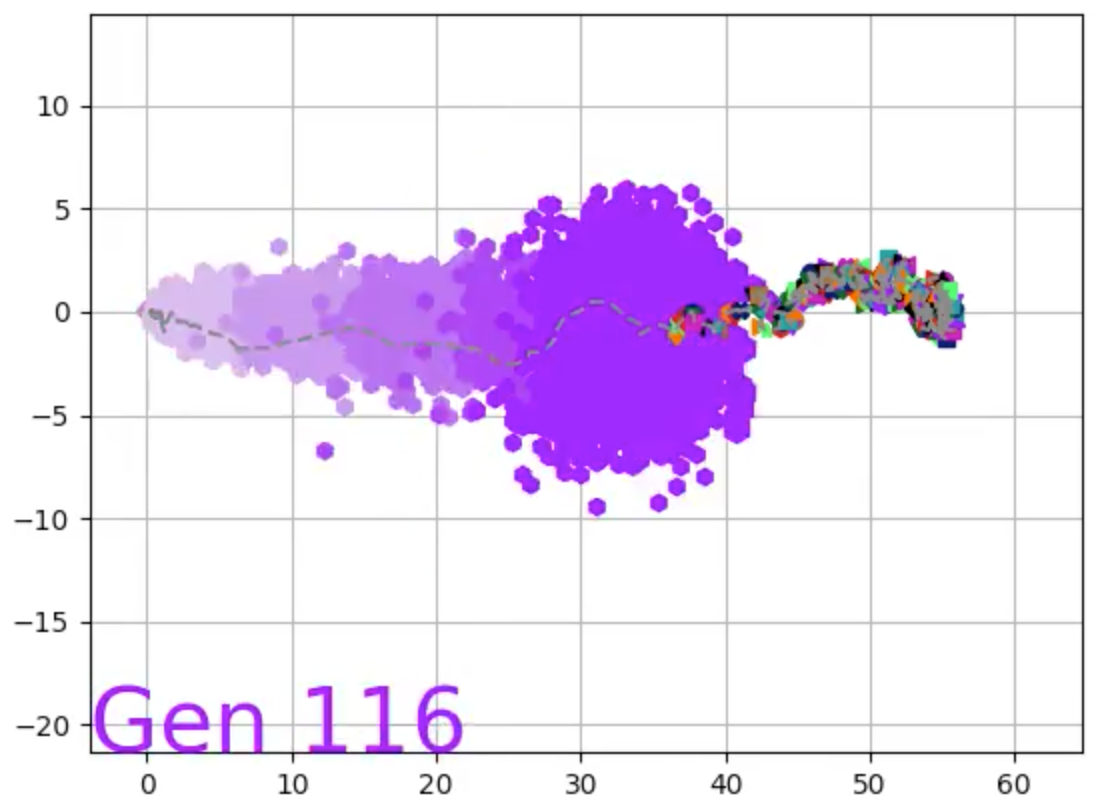}
        \caption{Video snapshot at Generation 116}
    \end{subfigure}\\%
  \begin{subfigure}[t]{0.5\textwidth}
        \centering
  \includegraphics[width=3.00in]{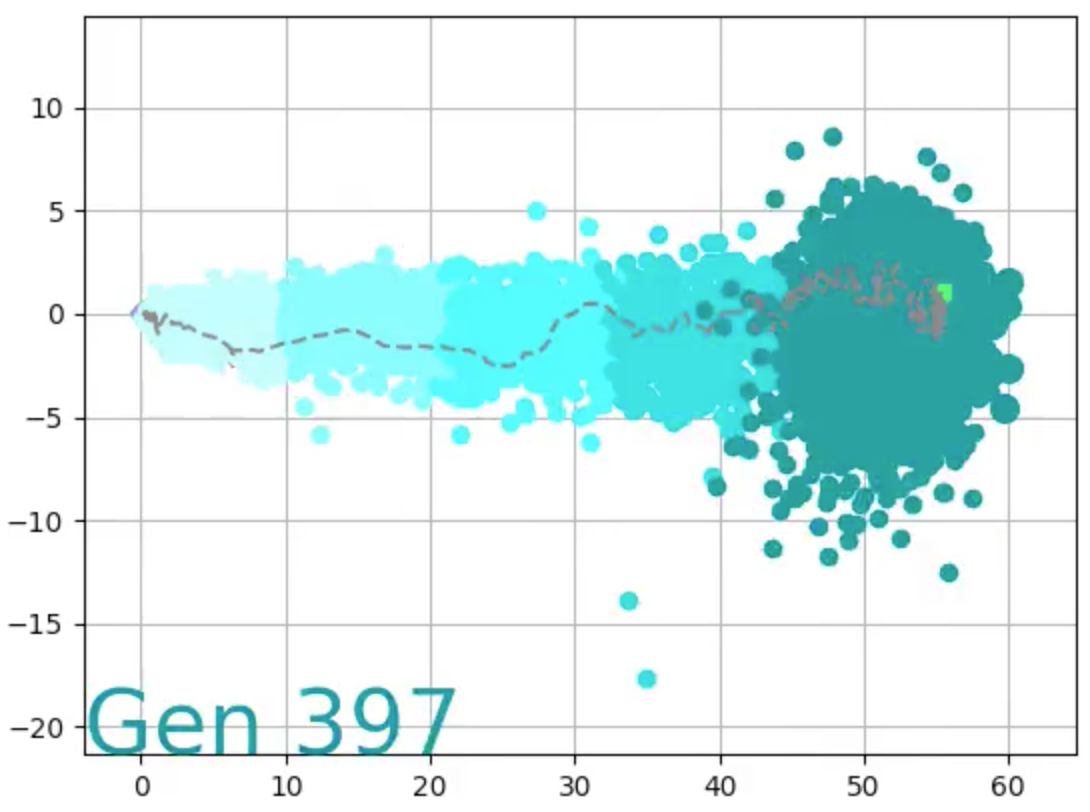}
  \caption{Video snapshot at Generation 397}
  \end{subfigure}
  \centering
	\vspace{-0.1in}
	\caption{\label{fig:cloud_movie}\textbf{Frames taken from a VINE-generated video visualizing the evolution of behaviors over generations in Humanoid Walking.} The color changes in each generation. Within a generation, the color intensity of each pseudo-offspring is based on the percentile of its fitness score in that generation (aggregated into five bins).  The position of each point corresponds to the endpoint of an individual walker (which was the BC in this example).}
\end{figure}

Users then interact with these plots to explore the overall trend of the pseudo-offspring cloud as well as the individual behaviors of any parent or pseudo-offspring over the evolutionary process: (1) users can visualize parents, top performers, and/or the entire pseudo-offspring cloud of any given generation, and explore the quantitative and spatial distribution on the 2D BC plane of pseudo-offspring with different fitness scores; (2) users can compare between generations, navigate through generations to visualize how the parent and/or the pseudo-offspring cloud is moving on the 2D BC plane, and how such moves relate to the fitness score curve (as illustrated in Figure \ref{fig:cloud_movie}, a full movie clip of the moving cloud can be generated automatically); (3) clicking on any point on the cloud plot reveals behavioral information and the fitness score of the corresponding pseudo-offspring.

\section{Additional use cases}


The tool also supports advanced options and customized visualizations beyond the default features. For example, instead of just a single final \{\emph{x}, \emph{y}\} point, the BC could instead be each agent’s full trajectory (e.g., the concatenated \{\emph{x}, \emph{y}\} for 1,000 time steps). In that case, where the dimensionality of the BC is above two, dimensionality reduction techniques (such as Principal Components Analysis (PCA) \citep{hotelling:pca33} or t-Distributed Stochastic Neighbor Embedding (t-SNE) \citep{Maaten:tsne08}) are needed to reduce the dimensionality of BC data to 2D. Our tool automates these dimensionality-reduction procedures. 

\begin{figure*}
  \centering
  \includegraphics[width=5.00in]{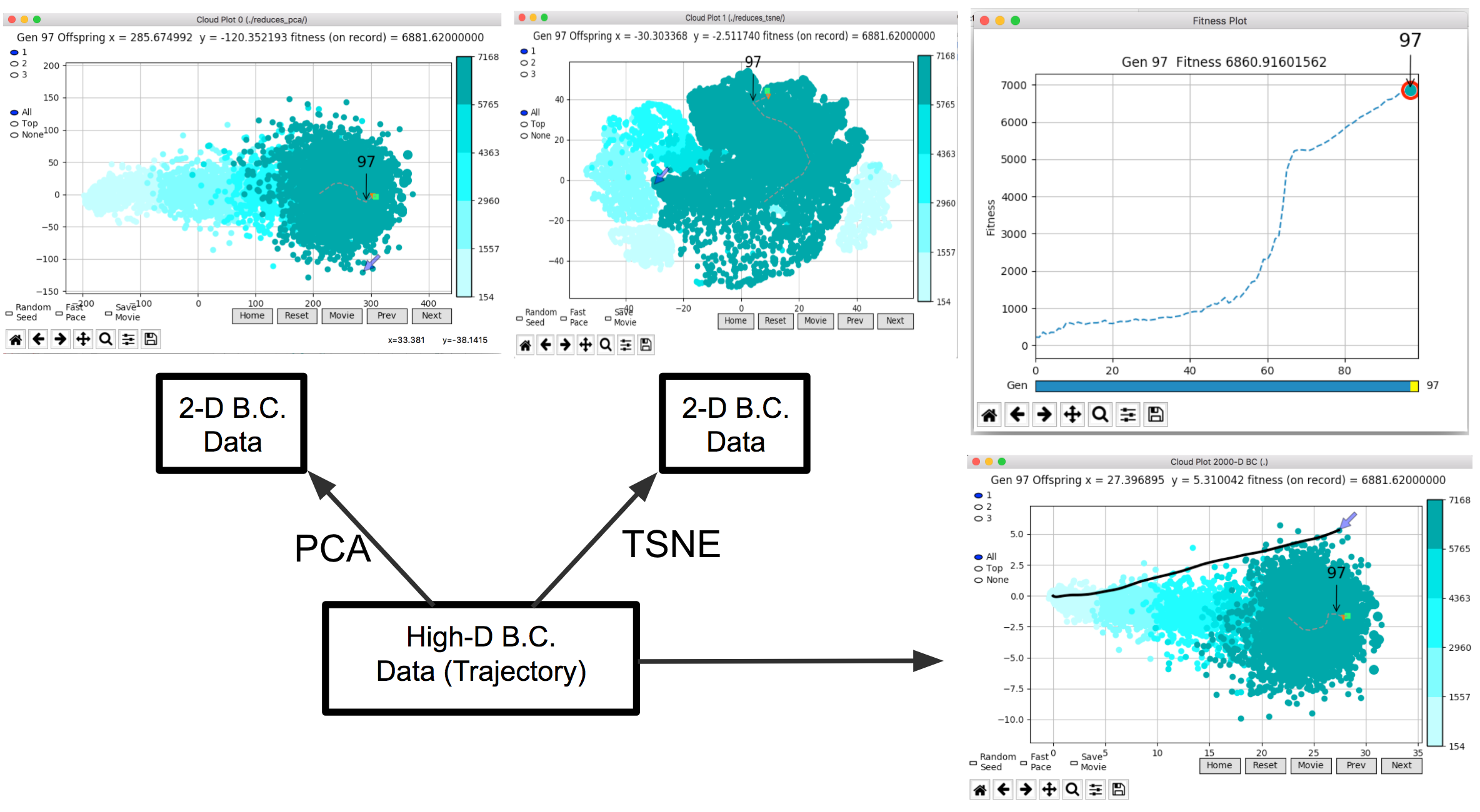}
  \caption{\label{fig:hidim_plot}\textbf{Visualizations of multiple 2D BCs and a high-dimensional BC along with a fitness plot.}  The three cloud plots show the same pseudo-offspring, but with their high-dimensional BCs reduced through different dimensionality reduction techniques, giving multiple perspectives on the space as it is searched.}
\end{figure*}

\begin{figure*}
  \centering
  \begin{subfigure}[t]{1.0\textwidth}
  		\centering
         \includegraphics[width=3.50in]{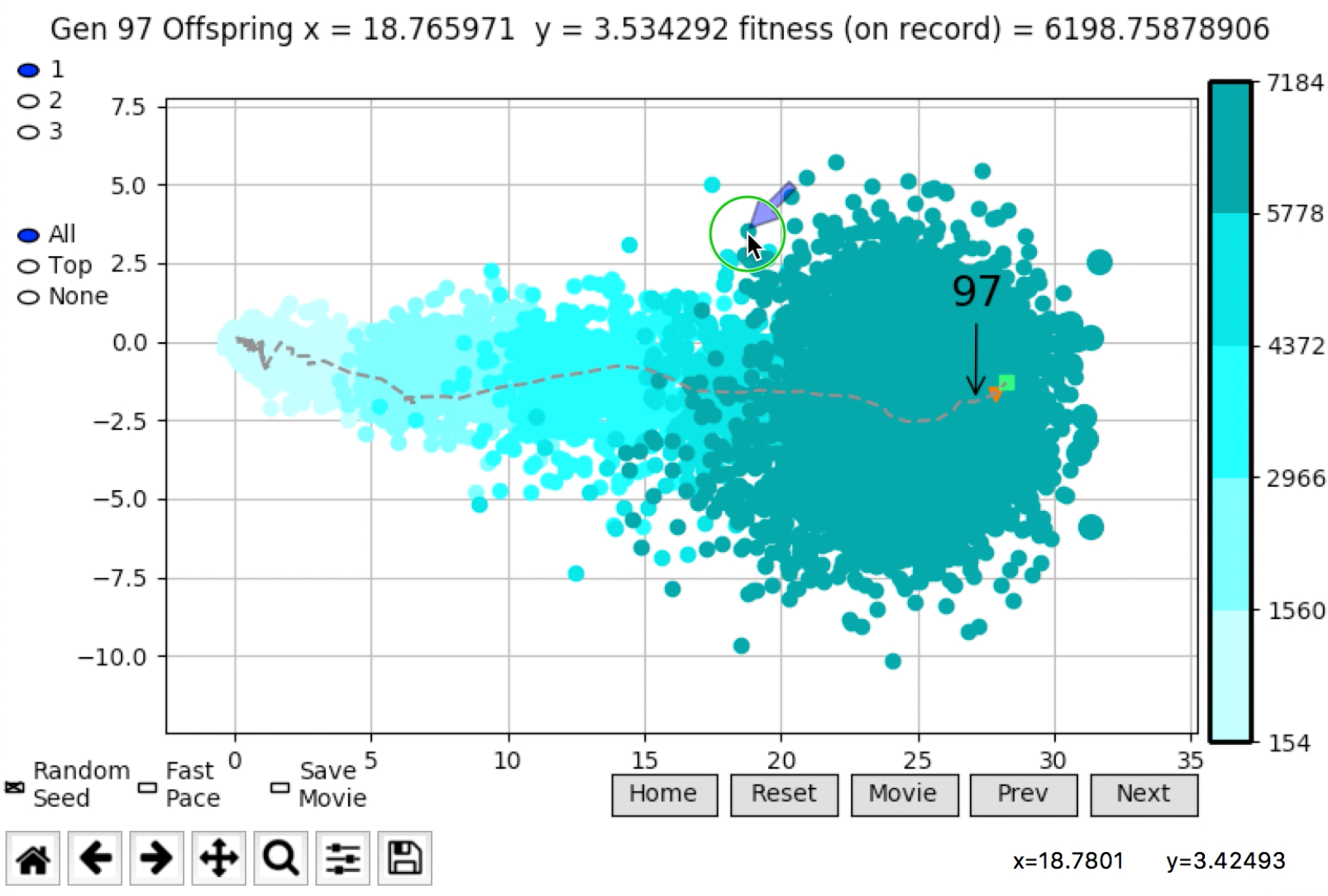}
        \vspace{-0.1in}
        \caption{Right click a pseudo-offspring to invoke nine stochastic roll-outs.}
    \end{subfigure}\\%
  \begin{subfigure}[t]{1.0\textwidth}
  		\centering
         \includegraphics[height=3.00in]{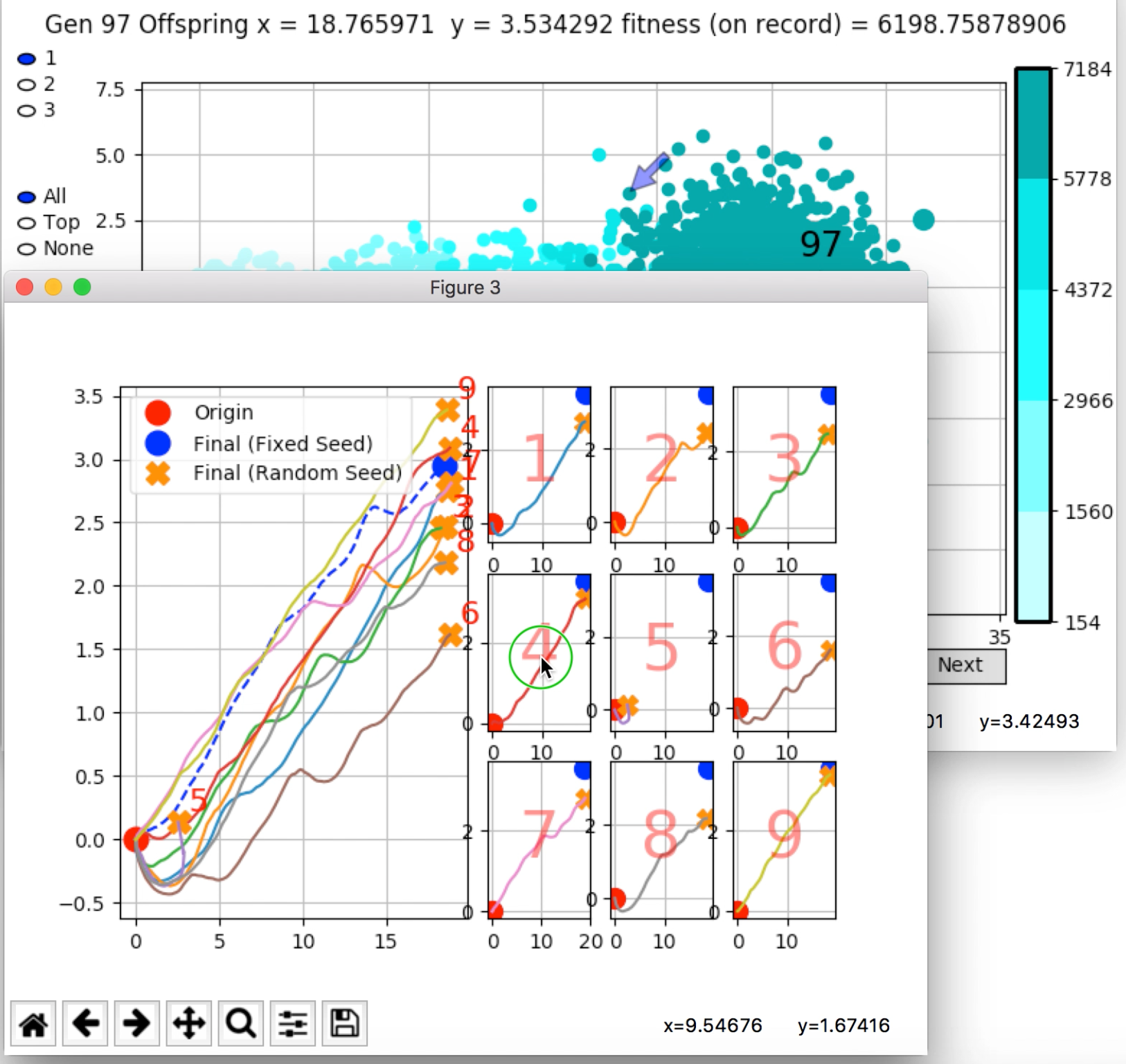}
        \vspace{-0.05in}
        \caption{Right click one of the trajectories as a result of nine roll-outs.}
    \end{subfigure}\\%
  \begin{subfigure}[t]{1.0\textwidth}
        \centering
  \includegraphics[width=3.50in]{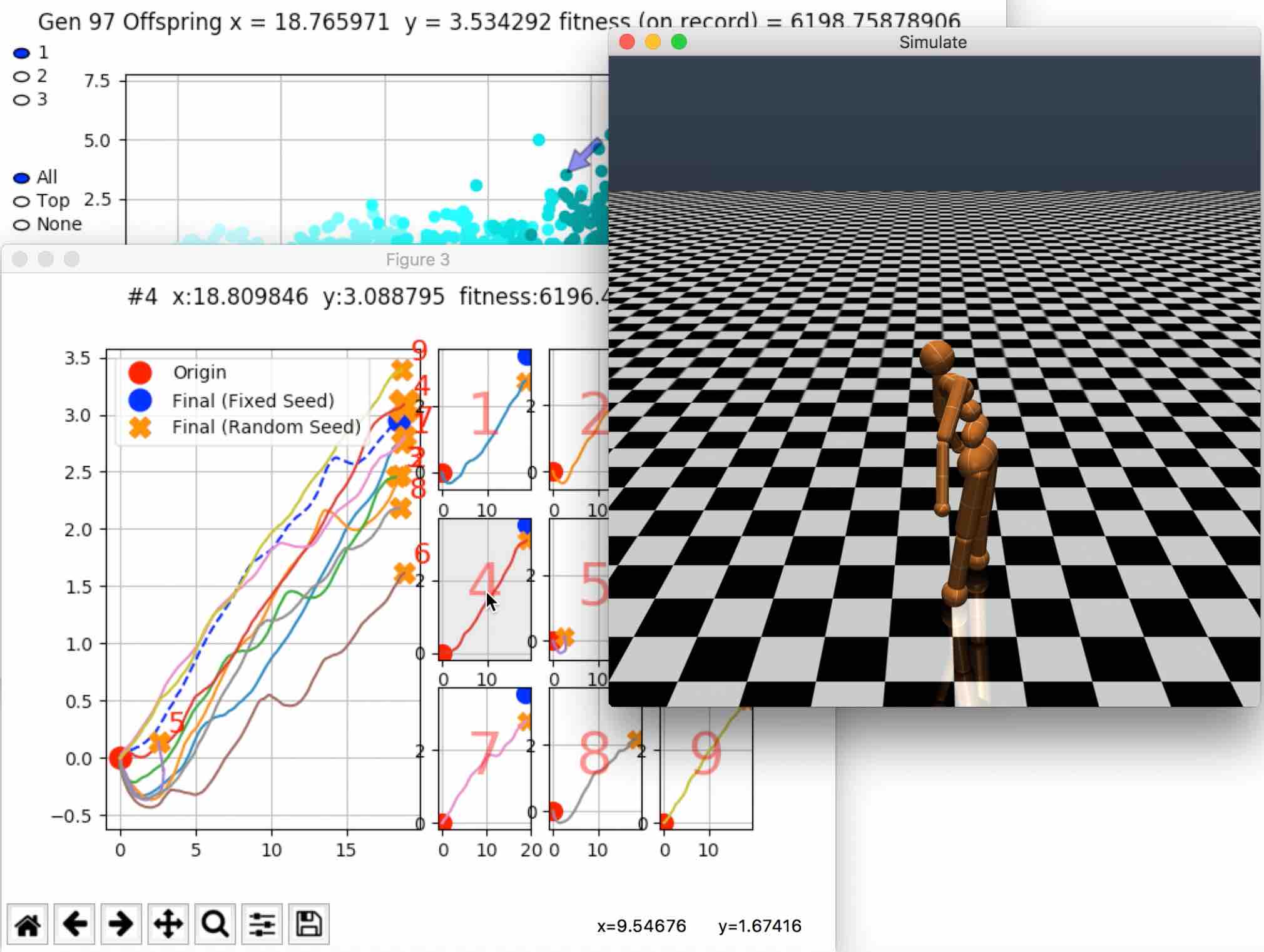}
  \vspace{-0.1in}
  \caption{Visualize the agent's behavior that corresponds to the trajectory in (b).}
  \end{subfigure}
  \centering
	\vspace{-0.1in}
	\caption{\label{fig:movie}\textbf{Users can view videos of any agent’s deterministic and stochastic behaviors through a video pop-up (at bottom).}}
\end{figure*}

The GUI is capable of loading multiple sets of 2D BCs (perhaps generated through different reduction techniques) and displaying them in simultaneous and connected cloud plots, as demonstrated in Figure \ref{fig:hidim_plot}. This capability provides a convenient way for users to explore different BC choices and dimensionality reduction methods. Furthermore, users can also extend the basic visualization with customized functionality. Figure \ref{fig:hidim_plot} exhibits one such customized cloud plot that can display certain types of domain-specific high-dimensional BCs (in this case, an agent’s full trajectory) together with the corresponding reduced 2D BCs. Another example of a customized cloud plot, in Figure \ref{fig:movie}, allows  the user to replay the agent’s deterministic or stochastic behavior that results when it interacts with an environment.

\begin{figure*}
  \centering
  \includegraphics[width=5.00in]{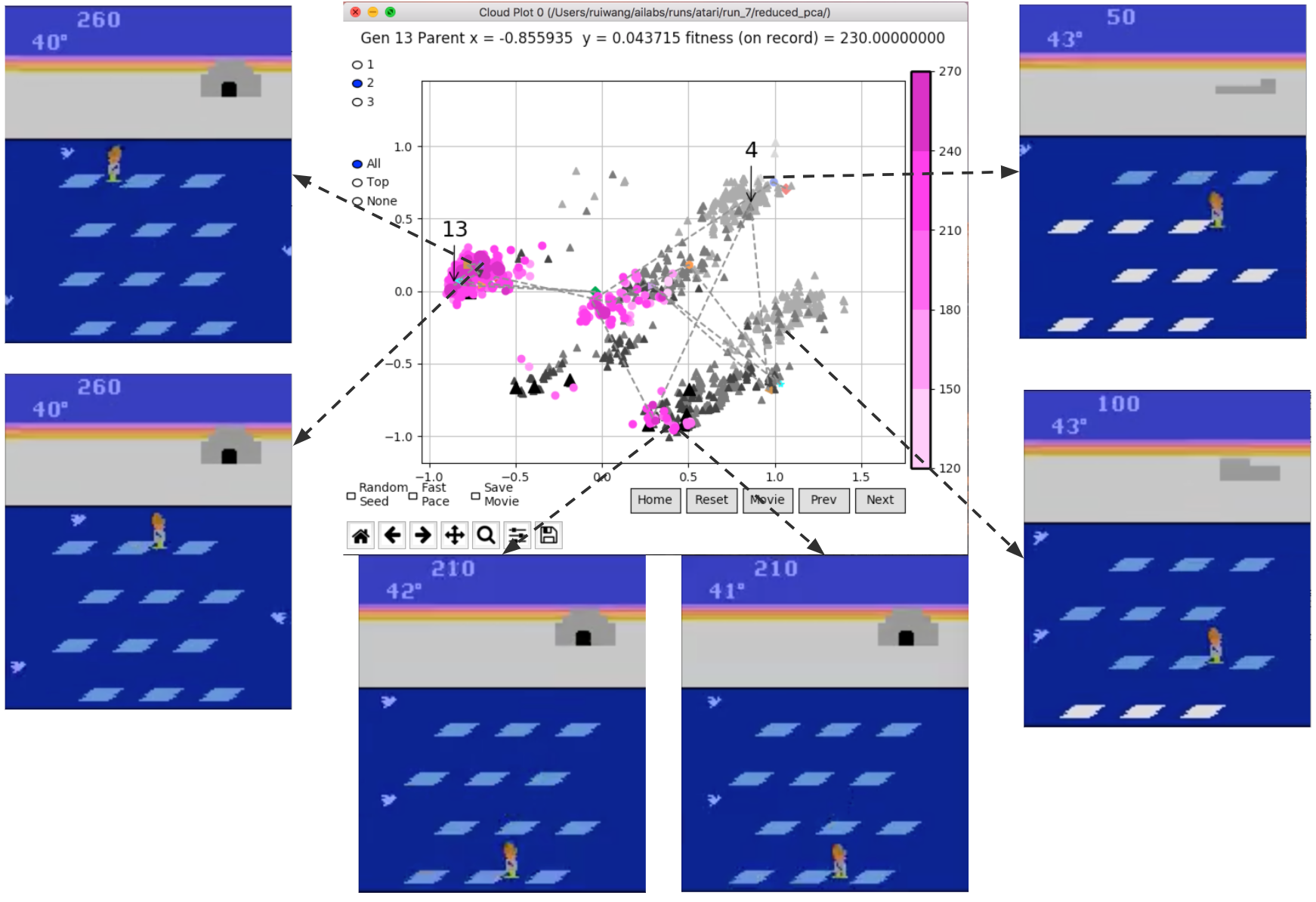}
  \caption{\label{fig:atari_plot}\textbf{Visualizing agents learning to play Frostbite.}  Each point is a 2D reduction of a high-dimensional representation of the end-state of the game for a particular psuedo-offpsring.  Users can click on any point to see the rollout of the game that leads to this endpoint, revealing the underlying semantics of the space.}
\end{figure*}

The tool is also designed to work with domains other than locomotion tasks. Figure \ref{fig:atari_plot} demonstrates a cloud plot that visualizes ES agents trained to play Frostbite, one of the Atari 2600 games \citep{Bellemare:atari13}, where we use the final emulator RAM state (integer-valued vectors of length 128 that capture all the state variables in a game) as the BC and apply PCA to map the BC onto a 2D plane.

The plot shows that as evolution progresses, the pseudo-offspring cloud shifts towards the left and clusters there. The ability to see the corresponding video of each of these agents playing the game lets us infer that each cluster corresponds to semantically meaningful and distinct end states.

VINE also works seamlessly with other neuroevolution algorithms such as GAs, which maintain a population of offspring over generations. In fact, the tool works independently of any specific neuroevolution algorithm. Users only need to slightly modify their neuroevolution code to save the BCs they pick for their specific problems. In the code release, we provide such modifications to our ES and GA implementations as examples.


\section{Conclusion}

Because evolutionary methods operate over a set of points, they present an opportunity for new types of visualization. Having implemented a tool that provides visualizations we found useful, we wanted to share it with the machine learning community so all can benefit. As neuroevolution scales to neural networks with millions or more connections, gaining additional insight through tools like VINE is increasingly valuable and important for further progress. 

\section*{Acknowledgements} 
We thank Uber AI Labs, in particular Joel Lehman, Xingwen Zhang, Felipe Petroski Such, and Vashisht Madhavan for valuable suggestions and helpful discussions.


\bibliography{smog,nn,ucf,vine}
\bibliographystyle{apalike}

\end{document}